\documentclass[letterpaper, 10 pt, conference]{ieeeconf}  

\IEEEoverridecommandlockouts                              

\usepackage{times}
\usepackage[pdftex]{graphicx}
\usepackage{subfigure}
\usepackage{amsmath,amssymb,amsopn,amstext,amsfonts}
\usepackage{cancel}
\usepackage[space]{cite}
\usepackage{pdfsync}
\usepackage{balance}
\usepackage{color}
\usepackage{mathtools}
\usepackage{bm}
\usepackage{diagbox}
\usepackage{float}
\usepackage{epstopdf}
\usepackage{pifont}
\usepackage{fixltx2e}
\usepackage{multirow}
\usepackage{url}
\usepackage{verbatim}
\usepackage[linkcolor=black,citecolor=black,urlcolor=black,colorlinks=true]{hyperref}
\usepackage{soul,xcolor}
\usepackage{algorithm,algorithmic}
\usepackage{subfiles}
\usepackage{makecell}
\graphicspath{{./figure/}}
\DeclareGraphicsExtensions{.png,.jpg,.eps,.pdf,.jpeg}

\title{\LARGE \bf
Fast 3D Sparse Topological Skeleton Graph Generation \\ for Mobile Robot Global Planning
}

\author{Xinyi Chen, Boyu Zhou, Jiarong Lin, Yichen Zhang, Fu Zhang and Shaojie Shen%
\thanks{This work was supported by HKUST Postgraduate Studentship
and HDJI Lab. X. Chen, B. Zhou, Y. Zhang and S. Shen are with the Department of Electronic and Computer Engineering, Hong Kong University of Science and Technology, Hong Kong, China. {\tt\footnotesize $\{$xchencq, bzhouai, eeshaojie$\}$@connect.ust.hk}. J. Lin and F. Zhang are with the Department of Mechanical Engineering, The University of Hong Kong, Hong Kong SAR, China. {\tt\small $\{$jiarong.lin, fuzhang$\}$@hku.hk}}%
}

\begin{document}
\maketitle
\thispagestyle{empty}
\pagestyle{empty}

\begin{abstract}
In recent years, mobile robots are becoming ambitious and deployed in large-scale scenarios.
Serving as a high-level understanding of environments, a sparse skeleton graph is beneficial for more efficient global planning. 
Currently, existing solutions for skeleton graph generation suffer from several major limitations, including poor adaptiveness to different map representations, dependency on robot inspection trajectories and high computational overhead.
In this paper, we propose an efficient and flexible algorithm generating a trajectory-independent 3D sparse topological skeleton graph capturing the spatial structure of the free space.
In our method, an efficient ray sampling and validating mechanism are adopted to find distinctive free space regions, which contributes to skeleton graph vertices, with traversability between adjacent vertices as edges.
A cycle formation scheme is also utilized to maintain skeleton graph compactness.  
Benchmark comparison with state-of-the-art works demonstrates that our approach generates sparse graphs in a substantially shorter time, giving high-quality global planning paths. 
Experiments conducted in real-world maps further validate the capability of our method in real-world scenarios.
Our method will be made open source to benefit the community.

\end{abstract}

\section{Introduction}
\label{sec:intro}

A sparse topological skeleton graph is a compact undirected graph structure capturing the spatial structure of the environment representing free space regions as vertices and their traversability as edges.
By providing a high-level understanding and abstraction of the environment, the sparse topological skeleton graph allows highly efficient global path planning, which is a fundamental problem for mobile robots.
Instead of finding paths directly on a map, which is computationally demanding especially in large-scale space, high-quality paths can be searched rapidly leveraging the skeleton graph.

Although a few methods have been developed to extract sparse 3D topological skeleton graphs, there are still some major limitations.
First, many of them need to pre-proceed the mapping results into the specific map representation that the methods are tailored for.
For example, \cite{oleynikova2018sparse} requires maintaining a Euclidean Signed Distance Field (ESDF) to build a Generalized Voronoi Diagram (GVD) and \cite{blochliger2018topomap} needs a grids-based map to grow clusters by dilating to neighboring grids.
These requirements make them less flexible to be applied elsewhere and take extra processing time.
Second, methods like \cite{blochliger2018topomap} typically follow a robot inspection trajectory to construct the skeleton graph.
Generated in such a manner, the skeleton graph heavily depends on the trajectory, instead of reflecting the inherent structure of the environment. 
Lastly, current methods are known to be computationally expensive, which can consume unreasonable time for large-scale environments.

\begin{figure}[t]
	\begin{center}          
    {\includegraphics[width=0.99\columnwidth]{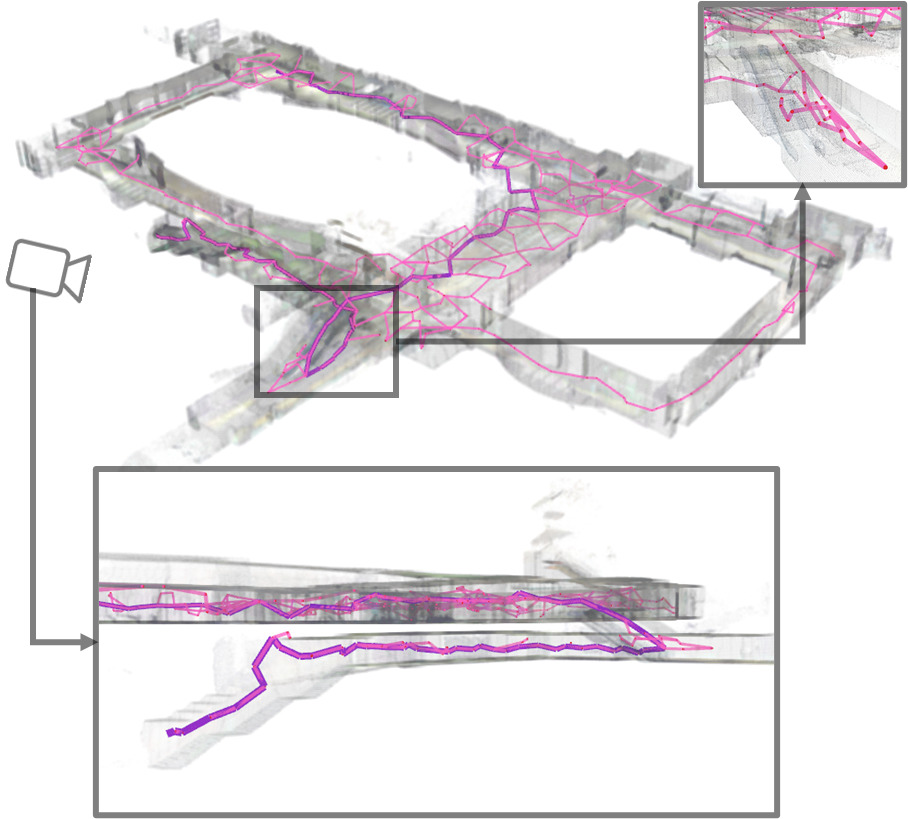}}       
  \end{center}
   \caption{\label{fig:realworld_large} A 3D sparse topological skeleton graph (pink) generated by the proposed method in a multi-floored real-world environment of size $110 \times 60 \times 13$ m$^3$ is demonstrated. A side-view from the left showing the multi-floor structure is placed at the bottom. Details of the skeleton graph passing an escalator are further enlarged in the top-right corner. Besides, a global planning A* path searched on the skeleton graph is marked in purple. For more details, please check out the video.}
   \vspace{-1.0cm}
\end{figure}

To address the aforementioned limitations, this work proposes an efficient and flexible approach to extract trajectory-independent sparse topological skeleton graphs from known environment maps.
Our algorithm supports various types of map representation, as long as a collision checking interface is available.
An efficient ray sampling and validating mechanism is adopted to iteratively extract distinctive free space regions, which contributes to vertices in the skeleton graph.
The traversability between adjacent vertices corresponds to the skeleton graph edges.
A cycle formation scheme is also utilized to maintain graph compactness.

We compare our approach with state-of-the-art works and evaluate the global planning performance of the skeleton graphs generated by different methods.
Results show that the proposed method generates sparse skeleton graphs in a substantially shorter time, capturing the spatial structure of free space precisely and giving high-quality global planning paths rapidly.
Moreover, we conduct experiments on real-world maps containing significant noises.
It demonstrates that our method is capable of handling noisy, complex and large-scale real-world environments, such as the multi-floored hall shown in Fig. \ref{fig:realworld_large}.

In summary, the contributions of this work are:
\begin{enumerate}
  \item An efficient and flexible algorithm that generates sparse trajectory-independent skeleton graph for mobile robots global planning, supporting various types of map representation as long as a collision checking interface is available.
  \item Benchmark comparison shows that the efficiency and skeleton graph quality of the proposed method outperforms the state-of-the-art works. The methodology is also validated in complex and large-scale real-world maps. The source code of the proposed method will be made available to benefit the community.
\end{enumerate}

\section{Related Work}
\label{sec:related}

Skeleton graph generation in a given environment for mobile robot global planning is gaining increasing attention by the community in recent years. 
Generating skeleton graphs for 2D environments has been a well-studied topic.
For the 2D case, most of the methods start with a Euclidean Signed Distance Field (ESDF) and build the skeleton making use of the Voronoi Diagram \cite{thrun1998learning, kalra2009incremental, liu2015incremental}.
3D skeleton graph generation is more difficult and still an open problem.
Existing methods can be summarized as GVD-based and clustering-based methods. 

\subsection{GVD-based Methods}
\label{subs:related_GVD}
It is natural to extend well-studied 2D methods into the 3D case, building Generalized Voronoi Diagrams (GVDs) for the skeleton graph generation.
\cite{hoff2000interactive} developed a 3D-applicable algorithm to compute Voronoi diagram by combining Voronoi diagrams for 2D parallel slices, which are computed using graphics hardware.
\cite{foskey2001voronoi} combines \cite{hoff2000interactive} and probabilistic roadmaps\cite{kavraki1996probabilistic} enabling path planning directly on the GVD graph.
Moreover, several computer graphics literature studies on 3D shape skeletonization for object meshes \cite{tagliasacchi2012mean} and point clouds \cite{tagliasacchi2009curve}.
On this topic, \cite{tagliasacchi20163d} presents a complete and detailed state-of-the-art report. 
However, these methods only consider regular small-scale objects, and are not directly applicable to large-scale maps containing substantial noises under mobile robot applications.

A recent GVD-based work \cite{oleynikova2018sparse} inspired by these works attempts to adapt these methods to complex and noisy environments for mobile robot navigation.
They first generate a one-voxel-thick skeleton diagram from GVD and further extract a skeleton graph from the diagram. 
However, it still results in graphs containing numerous vertices and edges for noisy maps, affecting global planning efficiency. 
Besides, this method requires an ESDF as input, whose maintenance is time-consuming, especially for large-scale environments.

\begin{table}[t]
  \centering
  \caption{\label{tab:frontier_attributes} Important attributes in a frontier}
  \begin{tabular}{cc}
    \hline\hline
    \textbf{Name}      & \textbf{Explanation}                              \\
    \hline
    Facets             & Component facets of the frontier                  \\
    Normal             & Average outward unit normals of facets            \\
    Center             & A central position on one of its facets \\
    Initial position   & Position of a new node construction                     \\
    Parent node        & Node holding this frontier                        \\
    \hline\hline
  \end{tabular}
\end{table}

\begin{table} [t]
  \centering
  \caption{\label{tab:vertex_attributes} Important attributes in a vertex}
  \begin{tabular}{cc}
    \hline\hline
    \textbf{Name}      & \textbf{Explanation}                         \\
    \hline
    Position           & Location of the vertex                       \\
    Type               & Black / White                                \\
    Detected polyhedron & The polyhedron that this vertex lies on \\
    Projected position & Vertex projection on the unit sphere        \\
    \hline\hline
  \end{tabular}
  \vspace{-0.2cm}
\end{table}

\subsection{Clustering-based Methods}
\label{subs:related_cluter}

The key idea of clustering-based methods is to divide an environment into meaningful clusters, after which skeletons are generated by connecting neighboring clusters. One way is to generate keyframe or landmark clusters based on a similarity measure in visual SLAM \cite{zivkovic2005hierarchical, blanco2006consistent, fraundorfer2007topological, vazquez2009spectral}. But landmark clusters capture no information of free space, which is not desirable for safe global path planning. Besides, redundant clusters representing the same place may be created when this place is visited from different directions. 
Another kind of approaches is attaching local occupancy sub-maps along the metric SLAM map \cite{konolige2011navigation}, where each sub-map form a cluster. 
These approaches only partially capture the topology of the space and can not be directly used for global planning.

A group of approaches that has advantages for motion planning is representing the environment as convex free-space clusters, as presented by some aerial robot trajectory planning literature \cite{deits2015computing,liu2017planning, zhong2020generating, chen2016online, gao2016online}. Along an initial path, a series of convex free-space clusters are built in shapes of convex polyhedra \cite{deits2015computing,liu2017planning, zhong2020generating}, cubes \cite{chen2016online} or spheres \cite{gao2016online} and further used in trajectory optimization. 
However, these methods only grow clusters for regions around the initial path, which can not be directly applied to construct sparse topological graphs of the entire environment. 
The most relevant work to ours is Topomap \cite{blochliger2018topomap}, which grows clusters along the explorer trajectory in an occupancy map. 
Convexity and compactness of each cluster are ensured by exhaustively checking all voxels in every growing iteration, costing an unreasonable amount of time for growing large clusters.

A common downside of these works is that they are highly dependent on robot inspection trajectory, instead of reflecting the inherent topological structure of the free space.
Centers of the clusters are often initialized along the given trajectory so the clusters only cover free space reached by it.

In contrast to these methods, we propose a novel approach that efficiently generates trajectory-independent sparse skeleton graph that reflects the topological structure of the environment and supports various types of map representation.

\section{Problem Statement}
\label{sec:problem}
Given a bounded and known map \(\mathcal{M}\), we aim to extract a 3D trajectory-independent sparse skeleton graph \(G\) reflecting the topological structure of the environment for mobile robot navigation.
Hence, narrow and inaccessible regions are not expected to be included in the resulting graph.
Serving as a high level understanding of the environment, the skeleton graph should be able to benefit the global planning for robots.

\section{Methodology}
\label{sec:proposed}

\subsection{Data Structures and Terminologies}
\label{subsec:data}

\begin{figure}[t]
  \begin{center}
    {\includegraphics[width=0.99\columnwidth]{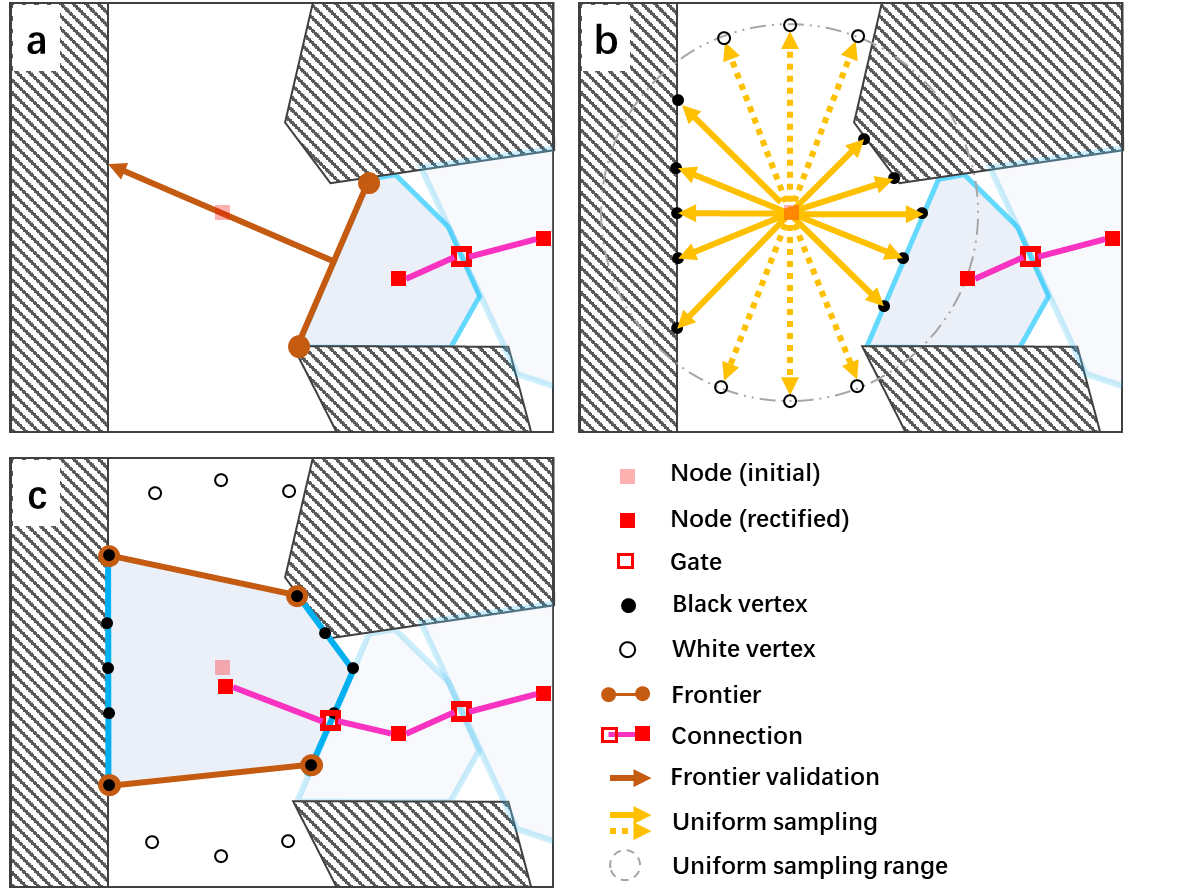}}
    \vspace{-0.8cm}
  \end{center}
  \caption{\label{fig:algworkflow} An illustration of the skeleton generation algorithm workflow. It shows an example algorithm iteration including frontier validation and node expansion.}
  \vspace{-1.2cm}
\end{figure}

To better explain the proposed approach, we would like to first introduce the data structures and some terminologies.
A \textit{node} represents a polyhedron covering a free space region and stores the polyhedron's facets and vertices.
In 2D cases (Fig. \ref{fig:algworkflow}), facets correspond to edges of the polygon.
Nodes also hold \textit{frontiers}, which are essentially special collections of adjacent facets.
A frontier stands on the border between its polyhedron and the free space that is not yet reached by any nodes, guiding the skeleton growth.
Important attributes stored in a frontier and vertex are listed in Table \ref{tab:frontier_attributes} and Table \ref{tab:vertex_attributes} respectively.
Besides, a \textit{gate} is built between two neighboring nodes and \textit{connections} are established between them indicating traversability.
In the resulting graph, both nodes and gates contribute to the vertices of the skeleton graph, whose edges are the connections.

\begin{algorithm}[t]
  \caption{Skeleton Graph Generation}
  \label{alg:workflow}
  \begin{algorithmic}[1]
    \renewcommand{\algorithmicrequire}{\textbf{Input:} Dense environment map $\mathcal{M}$}
    \renewcommand{\algorithmicensure}{\textbf{Output:} Sparse 3D skeleton graph \(G\)}
    \REQUIRE
    \ENSURE
    \STATE{$\mathcal{F}_{pndg}$, $\mathcal{N}$, $\mathcal{G}$, $\mathcal{C}$, $\mathcal{B}$  \(\leftarrow \emptyset\)  }
    \STATE{initialize($\mathcal{M}$, $\mathcal{F}_{pndg}$, $\mathcal{N}$, $\mathcal{B}$)}
    \WHILE{$\mathcal{F}_{pndg}$ not empty}
    \STATE{$f \leftarrow \mathcal{F}_{pndg}$.pop()}
    \STATE{verifyFrontier($f$)}
    \IF{$f$ invalid}
    \STATE{continue}
    \ENDIF
    \STATE{\(n \leftarrow\) new node($f.\text{initial\_position}$)} \label{alg1:node_expansion_start}
    \STATE{$\mathcal{V}_{black}, \mathcal{V}_{white} \leftarrow \emptyset $ }
    \STATE{generateVertices(\(n\), $\mathcal{V}_{black}$, $\mathcal{V}_{white}$, $\mathcal{B}$)}
    \IF{$\mathcal{V}_{white} = \emptyset$ \AND $n$.size $ \le \epsilon $}
    \STATE{continue}
    \ENDIF
    \STATE{cycleFormation($n, \mathcal{V}_{black}$)}
    \STATE{$P, \mathcal{F}_{new} \leftarrow$ buildPolyAndFrontier($n, \mathcal{V}_{black}$, $\mathcal{V}_{white}$)}
    \STATE{$\mathcal{B}$.push($\partial P$)}
    \STATE{$\mathcal{F}_{pndg}$.append($\mathcal{F}_{new}$)}
    \STATE{rectifyNodeCenter(\(n\), $\mathcal{V}_{black}$)} \label{alg1:node_expansion_end}
    \STATE{$g \leftarrow $ new gate(\(f\))}
    \STATE{$\mathcal{C}$.push(buildConnection($g, n$))}
    \STATE{$\mathcal{C}$.push(buildConnection($g, f.\text{parent\_node}$))}
    \ENDWHILE
    \RETURN  $ G \leftarrow (\mathcal{N} \cup \mathcal{G}, \mathcal{C})$
  \end{algorithmic}
\end{algorithm}

\subsection{Algorithm Overview}
\label{subsec:overview}
The algorithm workflow is described in Algorithm \ref{alg:workflow}.
At the beginning, the input map $\mathcal{M}$ is set and an initial node is expanded at an arbitrary position within free space.
During the initial node expansion, a few frontiers are identified and pushed into the pending frontiers First-In-First-Out(FIFO) list $\mathcal{F}_{pndg}$.
Then, the algorithm iteratively grows the skeleton graph in a breadth-first manner by popping out frontier $f$ from $\mathcal{F}_{pndg}$ and, if possible, expanding a new node.

In each iteration, $f$ is verified by performing raycasting from its center along its normal direction.
Only when the ray doesn't detect occupied space or polyhedron within a threshold distance from the center of $f$, it will be marked valid.
If $f$ is valid, the mid-point of the ray segment (brown arrow in Fig. \ref{fig:algworkflow}) is recorded in it.
Then, a new node $n$ is constructed taking this mid-point position as its initial position and a new gate $g$ is created at the center of $f$.
Next, $n$ is expanded in line \ref{alg1:node_expansion_start}-\ref{alg1:node_expansion_end}, which is the core of our algorithm (Section~\ref{subsec:node}).
During the node expansion, a polyhedron is grown while new frontiers are identified and pushed into $\mathcal{F}_{pndg}$ (Section \ref{subsec:frontiers}).
Moreover, skeleton graph cycles are formed if necessary (Section \ref{subsec:cycle}).
Upon expansion success, connections are built doubly-linked between $g$ and $n$, as well as $g$ and $f$'s parent node. The new nodes, gates and connections are collected in $\mathcal{N}$, $\mathcal{G}$  and $\mathcal{C}$ respectively.

\subsection{Node Expansion}
\label{subsec:node}
Starting from this section, we present the core of our algorithm: node expansion (line \ref{alg1:node_expansion_start}-\ref{alg1:node_expansion_end} in Algorithm \ref{alg:workflow}).
To expand a node $n$, we first perform raycasts from $n$'s initial position along uniformly sampled directions up to a truncated distance, as in Fig. \ref{fig:algworkflow}.
For each single ray, if no environment obstacles or other polyhedra is detected within the truncated distance, a white vertex $v_{white}$ is constructed at the truncated endpoint of the ray, indicating free space not yet reached by any node.
Otherwise, a black vertex $v_{black}$ is set at the first detected position on the ray segment.
If a ray detects a polyhedron, it indicates that there exists a cycle in the skeleton graph to be closed and the detected polyhedron is stored along with $v_{black}$.
To avoid growing redundant branches in the skeleton graph, a node holding no white vertex with a size smaller than a threshold $\epsilon$ is discarded, where the size of a node is computed as the mean distance from black vertices to the node center.
After passing the size check, a \textit{cycle formation} is performed (Section~\ref{subsec:cycle}) by iterating over all black vertices to close potential cycles in the skeleton graph.
Next, the polyhedron represented by $n$ is constructed, based on which new frontiers are identified (Section~\ref{subsec:frontiers}) and pushed to $\mathcal{F}_{pndg}$.
The boundary $\partial P$ of the newly grown polyhedron $P$ is recorded into $\mathcal{B}$, which will be used for raycasting for later nodes.
Finally, the position of $n$ is rectified as the average position of all $v_{black}$.



\begin{figure}[t]
  \begin{center}
    {\includegraphics[width=0.99\columnwidth]{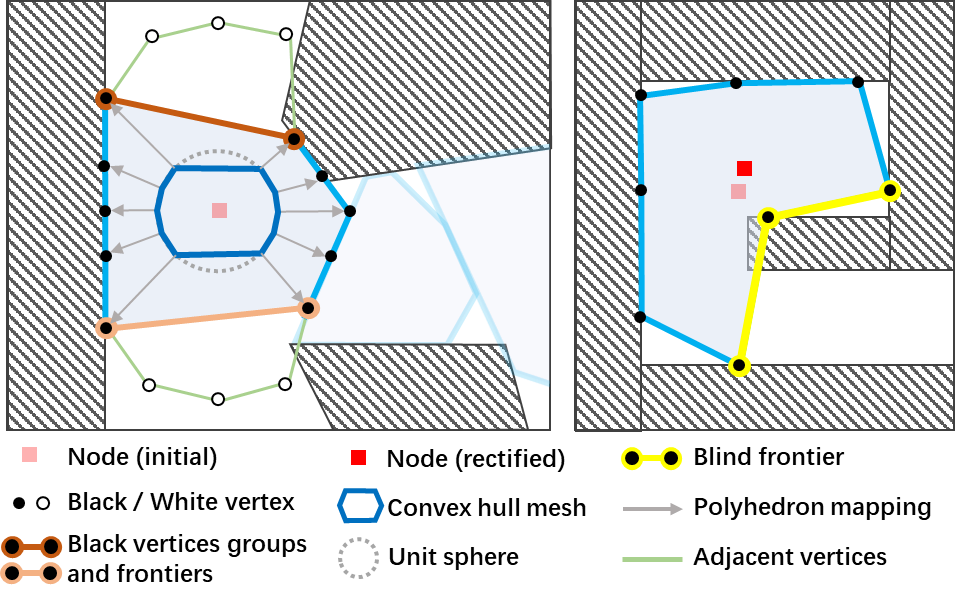}}
    \vspace{-0.8cm}
  \end{center}
  \caption{\label{fig:mesh} The left image illustrates the polyhedron construction and the right image is an example of the blind frontiers.
  }
   \vspace{-0.8cm}
\end{figure}

\subsection{Polyhedron Construction and Frontier Identification}
\label{subsec:frontiers}
Given black and white vertices list $\mathcal{V}_{black}$ and $\mathcal{V}_{white}$, we aim to construct the polyhedron $P$ of the current node $n$ and identify new frontiers $\mathcal{F}_{new}$ on the polyhedron boundary $\partial P$.
For all $v_{black}$ in $\mathcal{V}_{black}$, we first compute the projected positions $\hat{p}$ of $v_{black}$ on the unit sphere centered at $n$ as following:
$$\hat{p} = c + \frac{p-c}{||p-c||}$$
where $p$ is $v_{black}$ position and $c$ is the initial position of $n$.

To generate the polyhedron for $n$, a convex hull mesh $\hat{P}$ is computed on the projected positions $\hat{p}$ of all $v_{black}$.
Next, we map the facets of resulting mesh $\hat{P}$ back to the black vertices $\mathcal{V}_{black}$ to construct the polyhedron $P$, as in Fig. \ref{fig:mesh}.
Notice that the polyhedron constructed in this way is not necessarily convex and therefore is able to capture the free space more precisely.

To identify frontiers on $\partial P$, black vertices neighboring with adjacent white vertices are grouped as a set stored in $\mathcal{S}_{group}$.
Other black vertices not neighboring with any white vertex will remain ungrouped.
For a set $s_{i}$ of grouped black vertices, the frontier $f_{i}$ created by $s_{i}$ consists of all the polyhedron facets whose vertices all belong to $s_{i}$.
Then the new frontier $f_{i}$ will be split if the angle difference of its component facets' normals exceeds a threshold.
The split operation results in a few frontiers whose normals point towards different directions, guiding the skeleton graph to grow towards distinctive free space regions.
Moreover, a special kind of frontier, called \textit{blind frontier}, is identified to avoid missing possible passageway due to blind spot problems (Fig. \ref{fig:mesh}).
A blind frontier is formed by neighboring facets when their vertices have a large difference in distance to the node.
For each frontier, we calculate its normal as the average outward unit normals of its facets.
Also, the center of the frontier is computed as the projection of the average position of its facets centers along the normal to one of its facet. 
Finally, all the resulting frontiers $\mathcal{F}_{new}$ will be sorted decreasingly by the number of facets they own, since beyond a larger frontier there is a higher chance of discovering uncovered free space.

\begin{figure}[t]
  \begin{center}
    {\includegraphics[width=0.70\columnwidth]{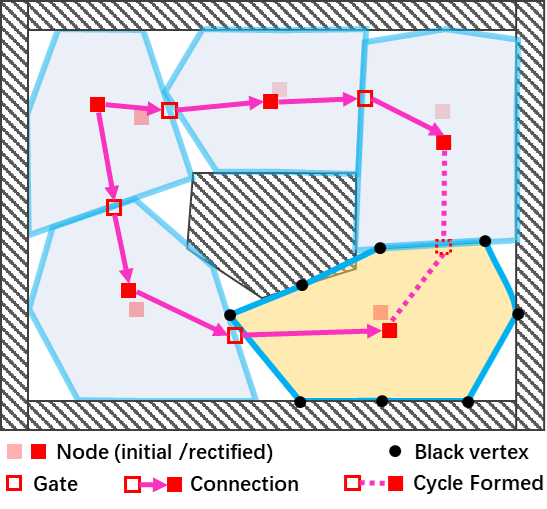}}
    \vspace{-0.5cm}
  \end{center}
  \caption{\label{fig:cycle} An illustration of a cycle formation, which builds connections and gate between the newly grown polyhedron (yellow) and a previous polyhedron. The skeleton graph generation process is indicated in arrows along with the connection segments.
  }
   \vspace{-0.2cm}
\end{figure}

\begin{figure}[t]
\begin{algorithm} [H]
  \caption{Polyhedron Construction/Frontier Identification}
  \label{alg:frontier}
  \begin{algorithmic}[1]
    \renewcommand{\algorithmicrequire}{\textbf{Input:} Black vertices $\mathcal{V}_{black}$, white vertices $\mathcal{V}_{white}$}
    \renewcommand{\algorithmicensure}{\textbf{Output:} Polyhedron $P$, new frontiers $\mathcal{F}_{new}$}
    \REQUIRE
    \ENSURE
    \STATE{$\mathcal{S}_{group}, \mathcal{F}_{new} \leftarrow \emptyset$}
    \FOR{$v_i \in \mathcal{V}_{black}$}
    \STATE{$\hat{p}_i \leftarrow$ computeProjectedPosition($v_i$)}
    \ENDFOR
    \STATE{$\hat{P} \leftarrow$ convexHullMesh($\{\hat{p}_i\}$)}
    \STATE{$P \leftarrow$ mapPolyhedron($\hat{P}$)}
    \STATE{$\mathcal{S}_{group} \leftarrow$ groupBlackVertices($\mathcal{V}_{black}$, $\mathcal{V}_{white}$)}
    \FOR{$s_i \in \mathcal{S}_{group}$}
    \STATE{$f_i \leftarrow$ createFrontier($s_i$)}
    \STATE{$\mathcal{F}_{new}$.append(splitFrontier($f_i$))}
    \ENDFOR
    \STATE{$\mathcal{F}_{new}$.append(blindFrontiers($\mathcal{V}_{black}$))}
    \STATE{sort($\mathcal{F}_{new}$)}
    \RETURN $P, \mathcal{F}_{new}$
  \end{algorithmic}
\end{algorithm}
\vspace{-1.5cm}
\end{figure}

\subsection{Cycle Formation}
\label{subsec:cycle}
A cycle is formed to ensure the compactness of the skeleton graph in the situation that the environment forms a loop, such as Fig. \ref{fig:cycle}.
To search cycles for the current node $n_{cur}$, we first find the polyhedron set $\{P_i, i \in \mathcal{I}\}$ whose elements are the detected polyhedron stored in all black vertices in $\mathcal{V}_{black}$.
Then, for each involved polyhedron $P_i$ and their corresponding node $n_i$ with $i\in \mathcal{I}$, we iterate through $n_i$'s frontiers and count the number of black vertices in $\mathcal{V}_{black}$ that lies on it.
The frontier $f^*$ with the most black vertices count takes part in the cycle formation.
A new gate $g$ is established at the center of $f^*$ and connections are set between $g$ and $n_{cur}$ as well as $g$ and $n_i$.
Then collision checks are performed on these two connections.
If any of them fails the check, the cycle formation will be revoked.

\section{Experimental Results}
\label{sec:results}
In this section, we evaluate the performance of our method in comparison with two representative state-of-the-art works. We also demonstrate our approach in real-world maps which contain significant noises.

\subsection{Implementation Details}
\label{subsec:implementation}

Our implementation of the proposed algorithm uses point cloud as an input map.
Hence, collision checking is performed by the nearest neighbor search provided by the \textit{Point Cloud Library (PCL)}\footnote{\url{https://www.pointclouds.org/}}.
The convex hull mesh computation used in Section~\ref{subsec:frontiers} is provided by \textit{quickhull}\footnote{\url{https://github.com/akuukka/quickhull.git}}.
All simulation experiments have been run in an Intel Core i7-4770K CPU at 3.5GHz, with 32 GB memory.

\begin{figure}[t]
   \centering
   \includegraphics[width=1.00\linewidth]{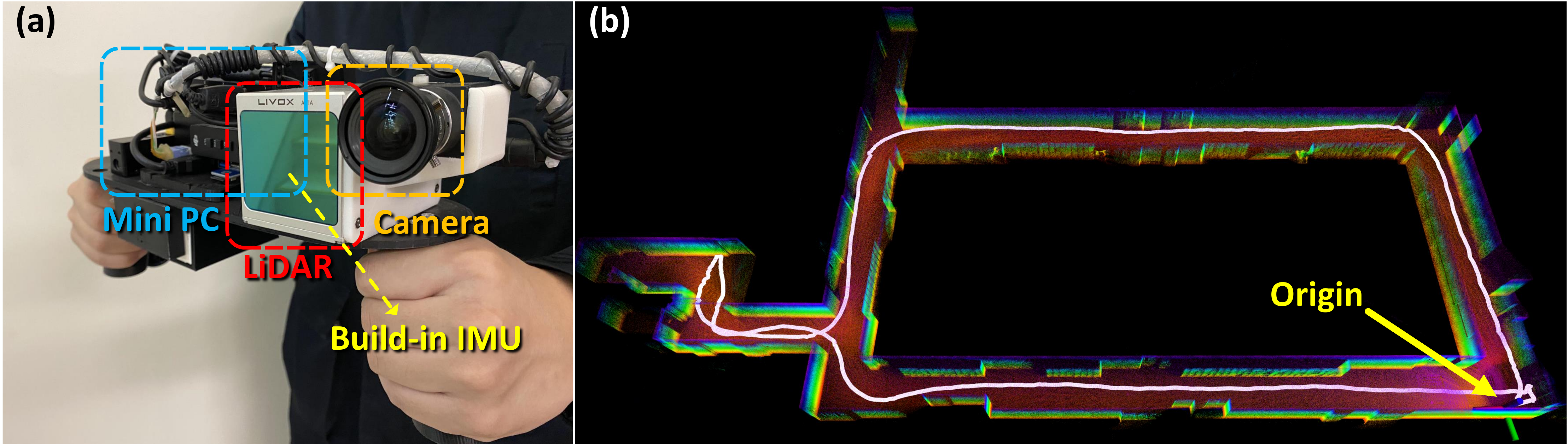}
   \caption{(a) Our handheld device for data collection. (b) The map is reconstructed by R$^3$LIVE, where the points are colored by their height and the white path is our traveling trajectory for sampling the data.}
   \label{fig:r3live_data_collection}
\end{figure}

\begin{table}[t]
   \centering
   \caption{\label{tab:benchmark1} Quality assessment of the skeleton graph}
      \begin{tabular}{>{\centering\arraybackslash}m{0.9cm}>{\centering\arraybackslash}m{0.8cm}cccc>{\centering\arraybackslash}m{0.5cm}>{\centering\arraybackslash}m{0.5cm}}
      \hline\hline
      \multirow{2}{*}{\textbf{Scene}}            & \multirow{2}{*}{\textbf{Method}} & \multicolumn{4}{c}{\textbf{Generation Time (s)}} & \multirow{2}{*}{\textbf{\#V}} & \multirow{2}{*}{\textbf{\#E}}                                                \\
                                                 &                                  & \textbf{Avg}                             & \textbf{Std}                  & \textbf{Max}                  & \textbf{Min}   &              &              \\
      \hline
      \multirow{3}{*}{Maze}                      & \cite{oleynikova2018sparse} & 69.35      & 0.278   & 69.78  & 68.79 & 7955 & 22015 \\
                                                 & \cite{blochliger2018topomap} &110.0         & 1.759   & 113.2  & 107.5 & \textbf{123} & \textbf{168}  \\
                                                 & Ours      & \textbf{2.126}  & 0.023                       &  2.180           & 2.100        & 464        & 484        \\
      \hline
      \multirow{3}{*}{\begin{tabular}[c]{@{}c@{}}Machine \\Hall\end{tabular}} &  \cite{oleynikova2018sparse}  & 1.029   & 1.081 & 0.977  & 0.417 & 531 & 1146 \\
                                                 & \cite{blochliger2018topomap} & 2.570  & 0.021  & 2.613  & 2.542 & \textbf{13} & \textbf{13} \\
                                                 & Ours     & \textbf{0.211} & 0.003                       & 0.219   & 0.208     & 69  & 77         \\
      \hline\hline
   \end{tabular}
   \vspace{-1.5cm}
\end{table}

\begin{figure}[t]
   \centering
   \vspace{-0.2cm}
   \subfigure{\includegraphics[width=0.48\columnwidth]{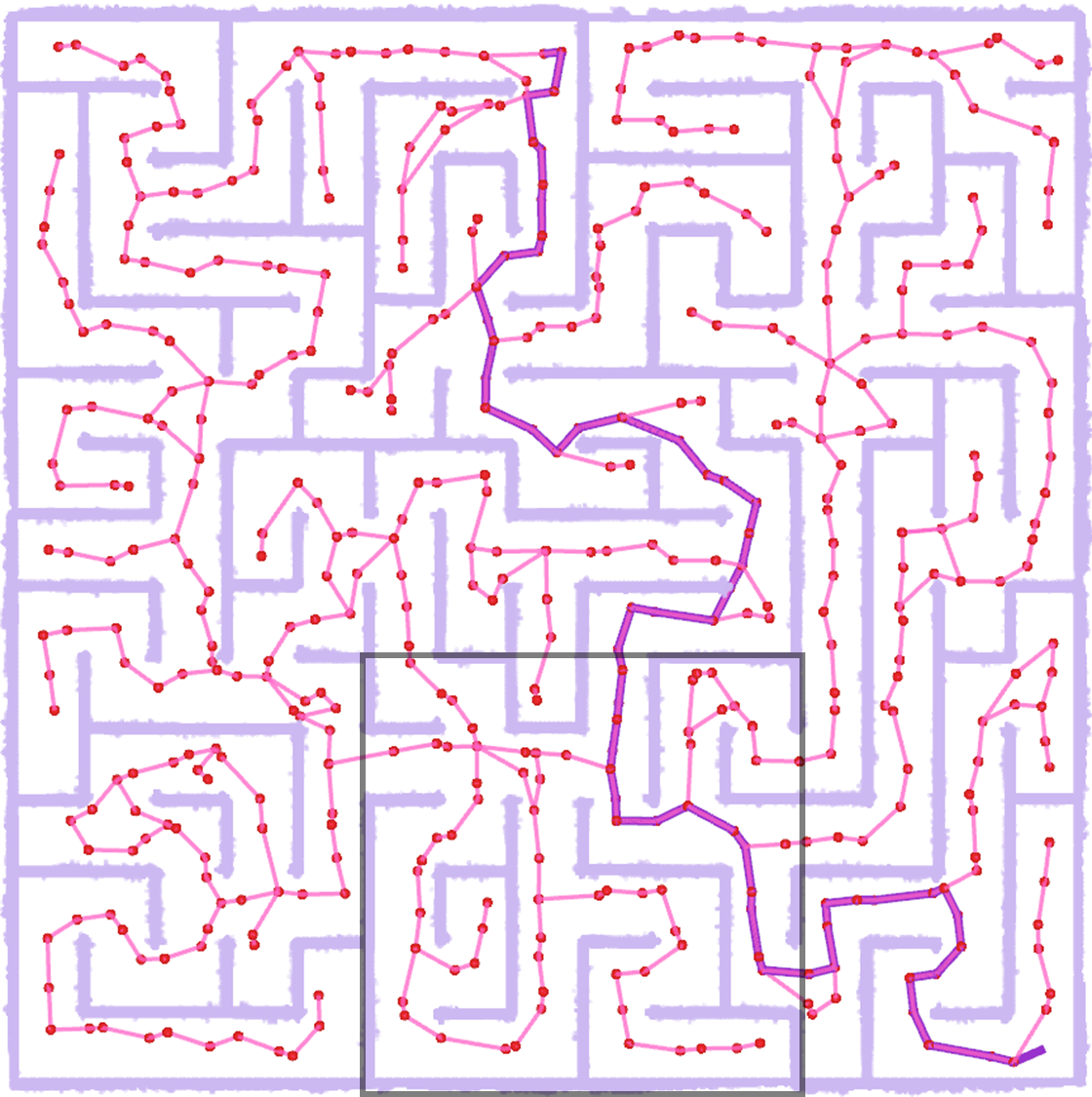}}
   \subfigure{\includegraphics[width=0.48\columnwidth]{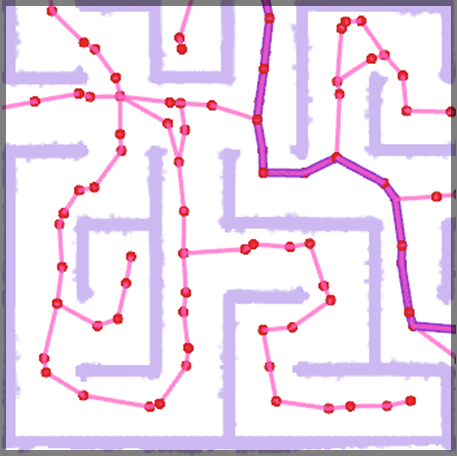}}
   \subfigure{\includegraphics[width=0.48\columnwidth]{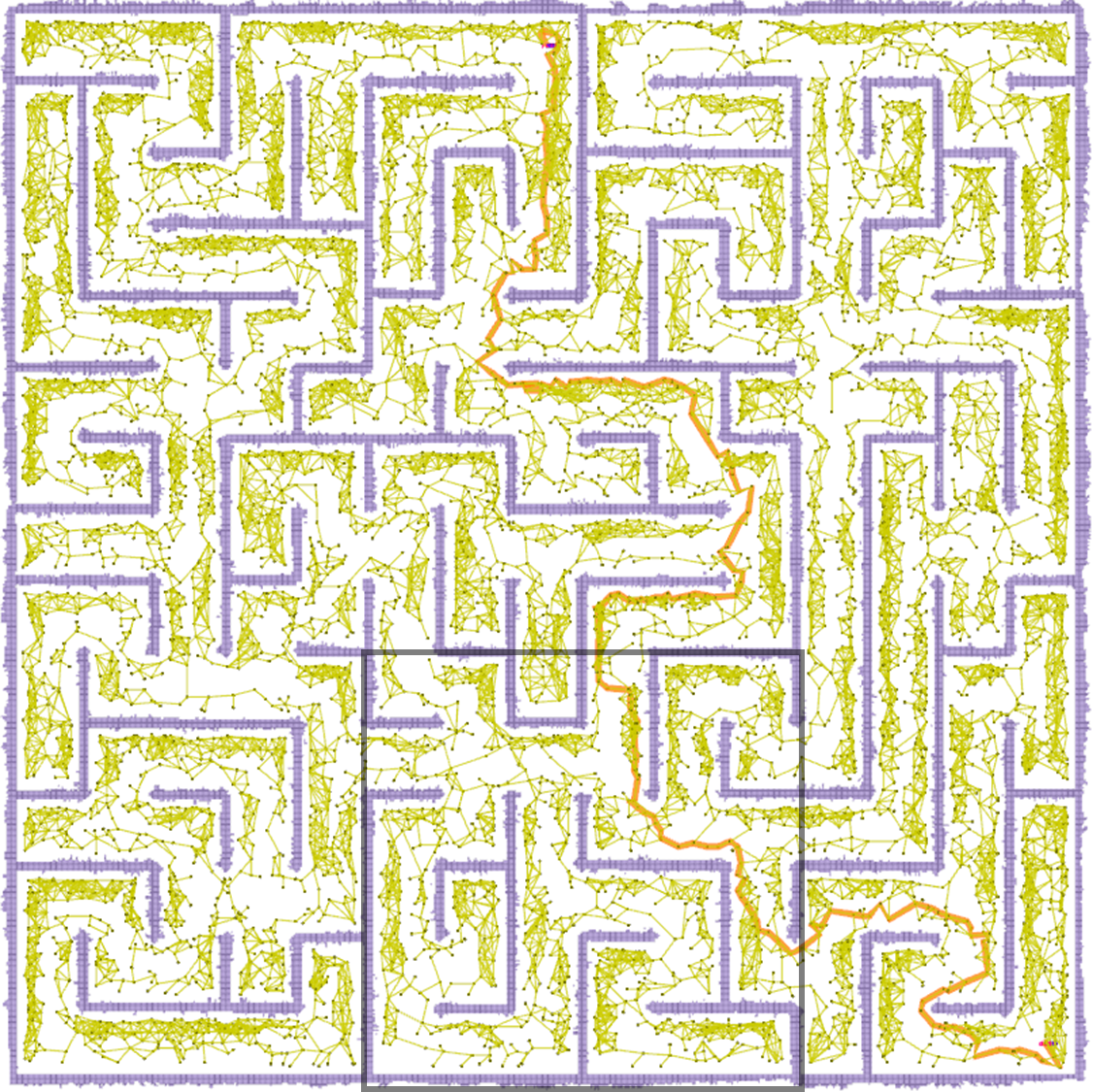}}
   \subfigure{\includegraphics[width=0.48\columnwidth]{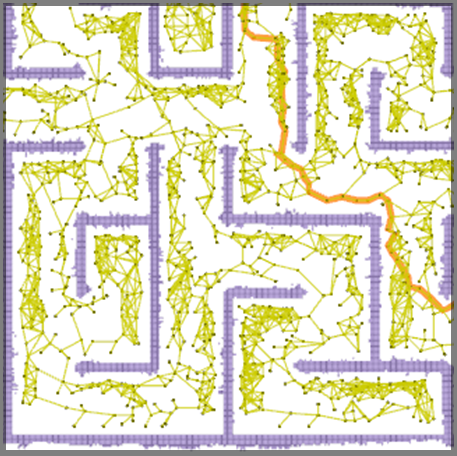}}
   \subfigure{\includegraphics[width=0.48\columnwidth]{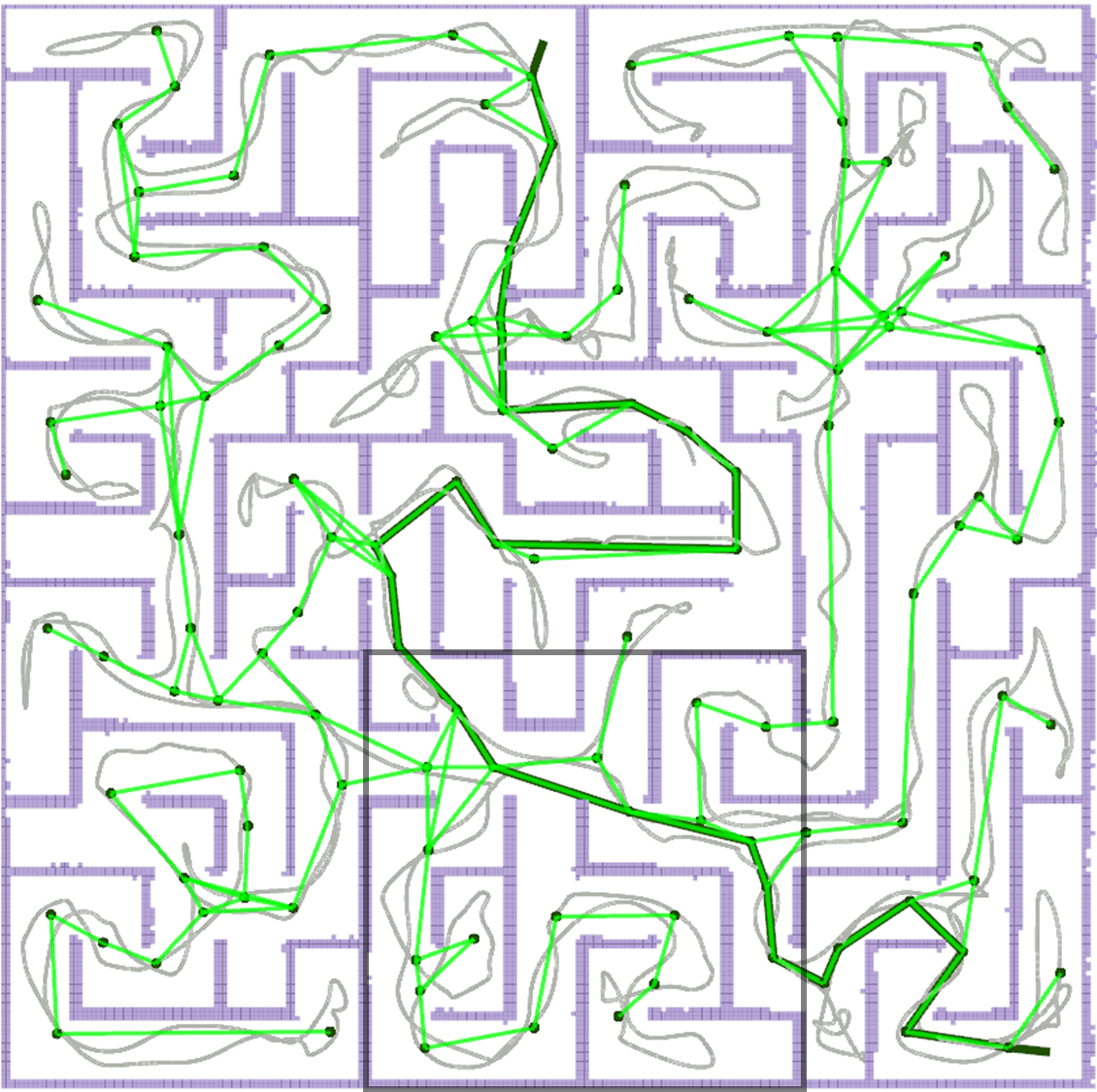}}
   \subfigure{\includegraphics[width=0.48\columnwidth]{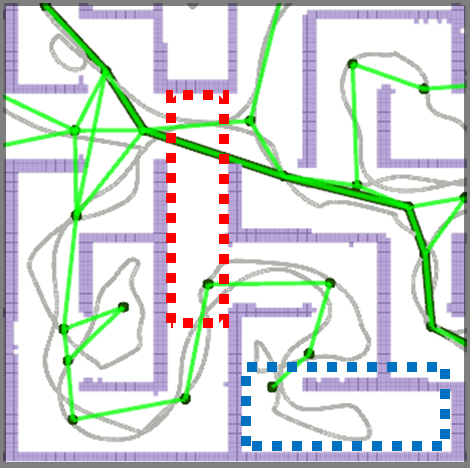}}
   \caption{Benchmark comparison results in the simulated maze scenario. From top to down, the works shown are proposed method,\cite{oleynikova2018sparse} and \cite{blochliger2018topomap}. Pictures in left column are the full views, with details in the black rectangles are shown in the right column. The bold line segments highlighted are the global planning path utilizing the skeleton graphs. The gray curve in the result of \cite{blochliger2018topomap} is the exploration trajectory used.}
   \label{fig:benchmark1}
   \vspace{-1.4cm}
\end{figure}

For real-world experiments, we collect the data with a handheld device shown in Fig. \ref{fig:r3live_data_collection}(a), which contains a mini PC \textit{DJI Manifold 2C}\footnote{\url{https://www.dji.com/manifold-2}}, a \textit{FLIR} global shutter camera,  and a \textit{LiVOX AVIA}\footnote{\url{https://www.livoxtech.com/avia}} LiDAR.
And to build the precise, dense, 3D point cloud of the surrounding environment in real-time, we leverage a low-drift, LiDAR-Visual-Inertial tightly-coupled state estimator R$^3$LIVE\cite{lin2021r3live} for reconstructing the maps (Fig. \ref{fig:r3live_data_collection}(b)).

\subsection{Benchmark Analysis}
\label{subs:results_benchmark}
In this section, we perform benchmark comparison with two state-of-the-art methods: a GVD-based method \cite{oleynikova2018sparse} and a clustering-based method \cite{blochliger2018topomap}.
Note that no open-source code is available for \cite{blochliger2018topomap} so we use our implementation.
For \cite{oleynikova2018sparse}, we transit point cloud maps into ESDFs, which is the required input map representation.
For \cite{blochliger2018topomap}, we perform an full coverage exploration utilizing \cite{zhou2021fuel} to obtain an exploration trajectory.
Also, an occupancy map with a voxel size of $0.25$m is built feeding as inputs together with the exploration trajectory.
The three methods are tested in two scenarios, for each of which, all methods are run 10 times with statistics shown in Table \ref{tab:benchmark1}.

\begin{table}[t]
   \centering
   \caption{\label{tab:benchmark2} Global planning performance}
   \begin{tabular}{ccccccc}
      \hline\hline
      \multirow{2}{*}{\textbf{Scene}}            & \multirow{2}{*}{\textbf{Method}} & \multicolumn{4}{c}{\textbf{A* Planning Time (ms)}} & \multirow{2}{*}{\textbf{Path(m)}}                                               \\
                                                 &                                  & \textbf{Avg}                             & \textbf{Std}                  & \textbf{Max}                  & \textbf{Min}   &              \\
      \hline
      \multirow{4}{*}{Maze}                      & Classical          & N/A  & N/A  & N/A  & N/A & N/A \\
                                                 & \cite{oleynikova2018sparse}    & 139.4   & 21.03 & 201.2 & 128.2 & 138.8 \\
                                                 & \cite{blochliger2018topomap}   & \textbf{0.729} & 0.066  &  0.818  & 0.596 & 122.4\\
                                                 & Ours       & 2.479   & 1.261    & 4.766      & 1.022     & \textbf{117.3}         \\
      \hline
      \multirow{4}{*}{\begin{tabular}[c]{@{}c@{}}Machine \\Hall\end{tabular}} & Classical & 5392 & 61.32    & 5460 & 5266 & \textbf{18.04} \\
                                                 & \cite{oleynikova2018sparse}         & 4.789 & 1.188 & 8.227  & 4.089 & 22.48 \\
                                                 & \cite{blochliger2018topomap}      & \textbf{0.055} & 0.011 & 0.071     & 0.035 & 42.19 \\
                                                 & Ours  & 0.434  & 0.133  & 0.712  & 0.265  & 21.74   \\

      \hline\hline
   \end{tabular}
\end{table}

\begin{figure}[t]
   \centering
   \subfigure{\includegraphics[height=0.6\columnwidth, angle = -90]{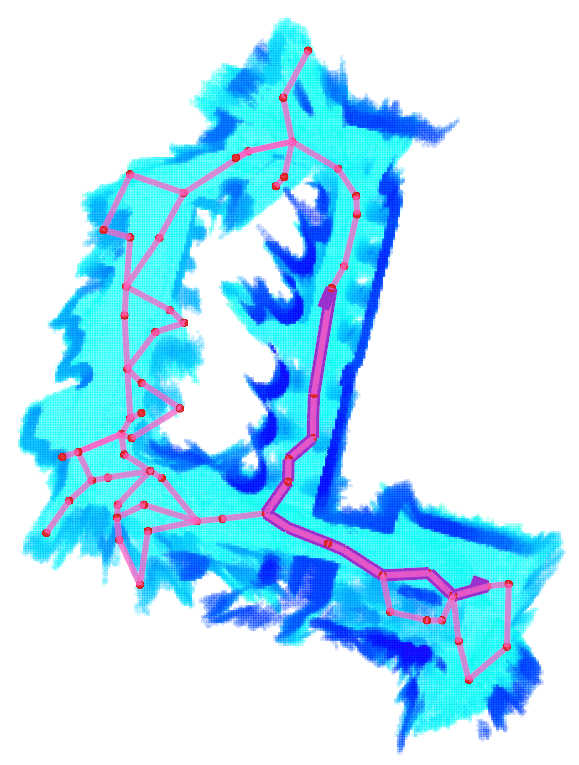}}
   \subfigure{\includegraphics[height=0.48\columnwidth, angle = -90]{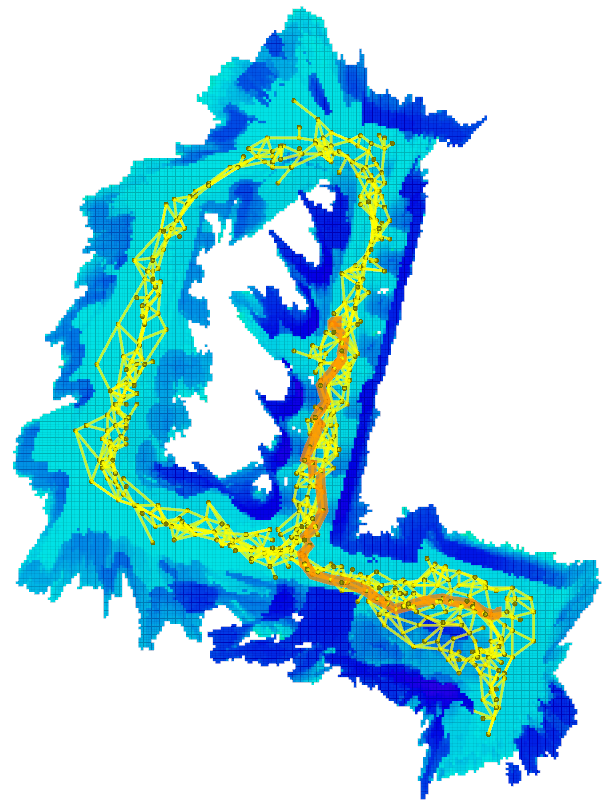}}
   \subfigure{\includegraphics[height=0.48\columnwidth, angle = -90]{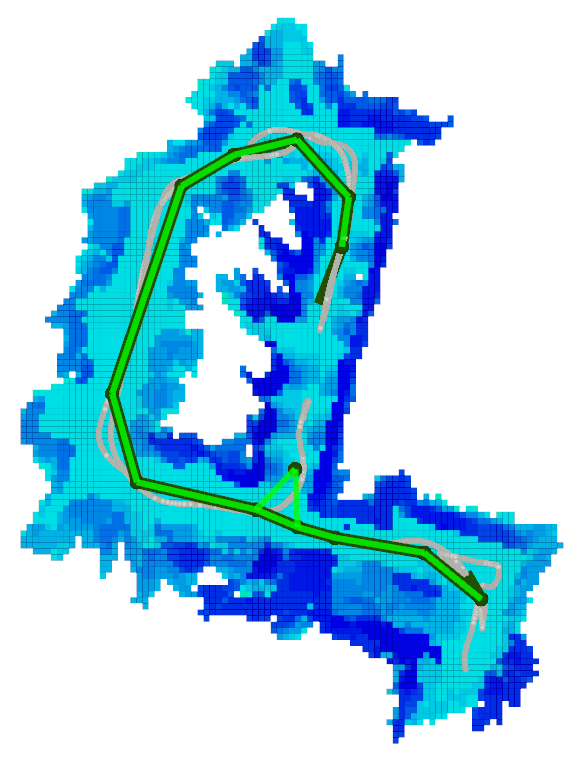}}
   \caption{Benchmark comparison results in the machine hall scenario. Our work is shown on top, with \cite{oleynikova2018sparse} and \cite{blochliger2018topomap} on bottom left and right respectively. The bold line segments highlighted are the global planning path utilizing the skeleton graphs. The gray curve in the bottom-right picture is the exploration trajectory used in \cite{blochliger2018topomap}. Note that the resulting skeleton graph from \cite{blochliger2018topomap} fails to include some important edges.}
   \vspace{-0.6cm}
   \label{fig:benchmark2}
\end{figure}

We first test the three methods in a $60 \times 60 \times 2.5$ m$^3$ simulated large maze scenario shown in Fig. \ref{fig:benchmark1}.
The results indicate that our method generates a sparse graph that reflects the topological structure of free space more precisely.
The skeleton graph generated by \cite{oleynikova2018sparse} is much denser, which has more than $17$ times vertices and $45$ times edges compared with the graph given by our method.
Although the skeleton graph produced by \cite{blochliger2018topomap} is sparse, it fails to capture the true structure of the free space since it grows along the exploration trajectory.
For example, in the bottom-right picture of Fig. \ref{fig:benchmark1}, the skeleton graph in red dashed rectangle should have been connected and it rejects to grow into dead ends such as the one in the blue dashed rectangle.
We also evaluate the three methods in a machine hall scenario, which is a $25 \times 30 \times 2.5$ m$^3$ open-sourse real-world dataset provided by \cite{oleynikova2018sparse}.
As shown in Fig. \ref{fig:benchmark2}, the experiments reveal that the three methods behave similarly as analyzed above and our method outperforms the others in skeleton graph quality.
In both scenarios, our method achieves a much shorter skeleton graph generation time compared with \cite{oleynikova2018sparse} and \cite{blochliger2018topomap}.
Especially in large-scale environments such as the maze scenario, our approach is $30+$ times faster than the other two methods.

Moreover, we evaluate of global planning performance utilizing the resulting skeleton graphs of the three methods by the A* algorithm.
To show the advantages of the skeleton graph, we also compare it with classical A* algorithm, which performs a search on grids.
The global planning paths are highlighted in bold line segments in Fig. \ref{fig:benchmark1} - \ref{fig:benchmark2} and statistics are shown in Table \ref{tab:benchmark2}.
Notice that the classical A* algorithm is not able to finish in $10$ minutes in the large maze scenario.
Global planning on the skeleton graphs generated by our method is at least $10$ times faster compared with \cite{oleynikova2018sparse} and gives the shortest path among the three skeleton graphs.
Although global planning on the skeleton graph given by \cite{blochliger2018topomap} is fast, it fails to give the optimal path topology in tasks for both scenarios because of the incompleteness of the skeleton graph.
Compared with classical global planning on grids, the A* algorithm utilizing skeleton graphs rendered by our method is able to give competitive high-quality paths $20,000$+ times faster in only a few milliseconds.

\subsection{Real-world Map Experiments}
\begin{figure}[t]
   \begin{center}
      {\includegraphics[width=0.9\columnwidth]{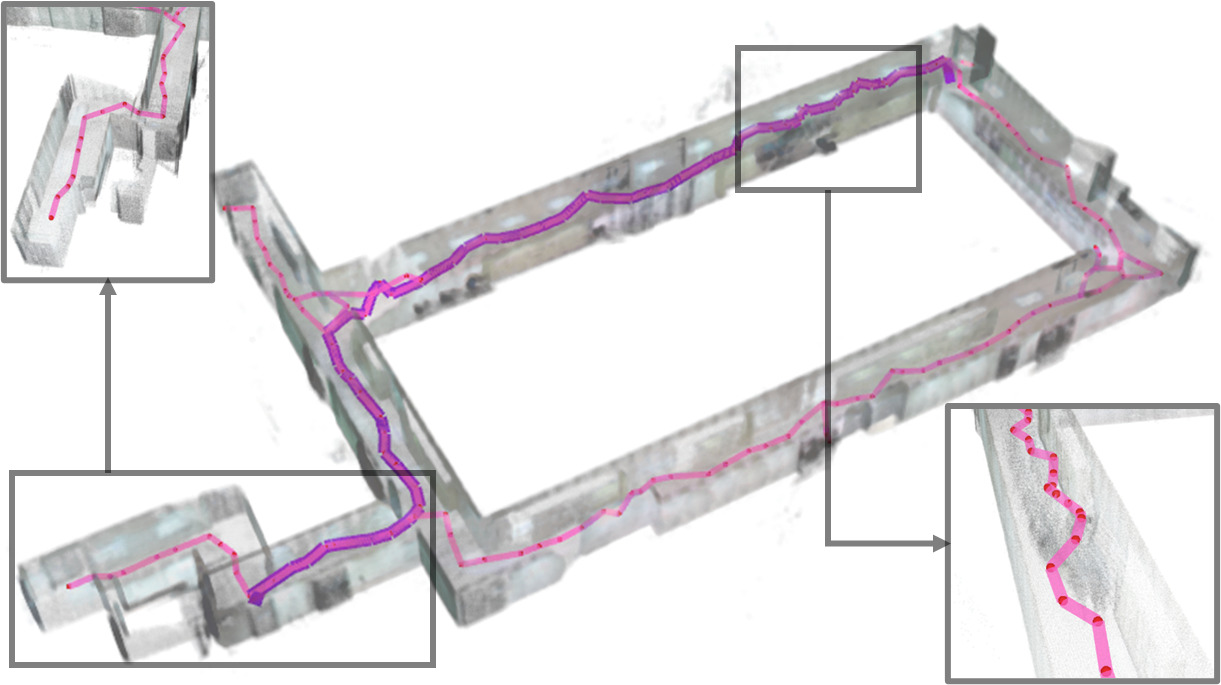}}
      \vspace{-0.2cm}
    \end{center}
   \caption{This is an indoor corridor of size $40 \times 30 \times 2$ m$^3$. Skeleton graph is shown in pink and the A* global planning path is highlighted in purple. Pictures at the top-left and bottom-right corners show the details.}
   \vspace{-0.8cm}
   \label{fig:realworld_simple}
\end{figure}

To further validate the feasibility of the proposed method, we conduct experiments on real-world data collected by us using the handheld device as mentioned in Sec. \ref{subsec:implementation}.
As in the enlarged picture at the bottom-right of Fig. \ref{fig:realworld_simple}, someone walked by and provided substantial noises to the map.
The skeleton graph with $162$ vertices and $167$ edges is successfully generated in $0.654$ seconds, demonstrating that our method is capable of handling noisy maps.
An A* global planning path with a length $58.69$m is searched in $2.087$ milliseconds.
For the large-scale multi-floored hall scenario in Fig.~\ref{fig:realworld_large}, the skeleton graph with $438$ vertices and $523$ edges is generated in $2.856$ seconds and the A* global planning path with length $178.1$m is searched in $3.788$ milliseconds.
The experiments further validate the capability of our algorithm in noisy, complex and large-scale maps.

\section{Conclusions}
\label{sec:conclude}
In this paper, we proposed an efficient and flexible approach to generate trajectory-independent sparse topological skeleton graphs for mobile robot global planning.
Our method adopts an efficient ray sampling and validating mechanism to iteratively extract polyhedra in free space.
Frontiers are identified on the boundary of polyhedra, guiding the skeleton graph to
grow towards distinctive free space regions.
Compared with state-of-the-art works, our method generates sparse skeleton graph in a substantially shorter time, capturing the spatial structure of free space precisely and giving high-quality global planning paths rapidly.
Experiments on real-world maps demonstrate our method is capable of handling noisy, complex and large-scale environments.


\addtolength{\textheight}{0.cm}   

\newlength{\bibitemsep}\setlength{\bibitemsep}{0.0\baselineskip}
\newlength{\bibparskip}\setlength{\bibparskip}{0.1pt}
\let\oldthebibliography\thebibliography
\renewcommand\thebibliography[1]{%
\oldthebibliography{#1}%
\setlength{\parskip}{\bibitemsep}%
\setlength{\itemsep}{\bibparskip}%
}

\bibliography{ref}

\begin{thebibliography}{10}
\providecommand{\url}[1]{#1}
\csname url@samestyle\endcsname
\providecommand{\newblock}{\relax}
\providecommand{\bibinfo}[2]{#2}
\providecommand{\BIBentrySTDinterwordspacing}{\spaceskip=0pt\relax}
\providecommand{\BIBentryALTinterwordstretchfactor}{4}
\providecommand{\BIBentryALTinterwordspacing}{\spaceskip=\fontdimen2\font plus
\BIBentryALTinterwordstretchfactor\fontdimen3\font minus
  \fontdimen4\font\relax}
\providecommand{\BIBforeignlanguage}[2]{{%
\expandafter\ifx\csname l@#1\endcsname\relax
\typeout{** WARNING: IEEEtran.bst: No hyphenation pattern has been}%
\typeout{** loaded for the language `#1'. Using the pattern for}%
\typeout{** the default language instead.}%
\else
\language=\csname l@#1\endcsname
\fi
#2}}
\providecommand{\BIBdecl}{\relax}
\BIBdecl

\bibitem{oleynikova2018sparse}
H.~Oleynikova, Z.~Taylor, R.~Siegwart, and J.~Nieto, ``Sparse 3d topological
  graphs for micro-aerial vehicle planning,'' in \emph{Proc. of the {IEEE/RSJ}
  Intl. Conf. on Intell. Robots and Syst.({IROS})}.\hskip 1em plus 0.5em minus
  0.4em\relax IEEE, 2018, pp. 1--9.

\bibitem{blochliger2018topomap}
F.~Blochliger, M.~Fehr, M.~Dymczyk, T.~Schneider, and R.~Siegwart, ``Topomap:
  Topological mapping and navigation based on visual slam maps,'' in \emph{2018
  IEEE International Conference on Robotics and Automation (ICRA)}.\hskip 1em
  plus 0.5em minus 0.4em\relax IEEE, 2018, pp. 1--9.

\bibitem{thrun1998learning}
S.~Thrun, ``Learning metric-topological maps for indoor mobile robot
  navigation,'' \emph{Artificial Intelligence}, vol.~99, no.~1, pp. 21--71,
  1998.

\bibitem{kalra2009incremental}
N.~Kalra, D.~Ferguson, and A.~Stentz, ``Incremental reconstruction of
  generalized voronoi diagrams on grids,'' \emph{Robotics and Autonomous
  Systems}, vol.~57, no.~2, pp. 123--128, 2009.

\bibitem{liu2015incremental}
M.~Liu, F.~Colas, L.~Oth, and R.~Siegwart, ``Incremental topological
  segmentation for semi-structured environments using discretized gvg,''
  \emph{Autonomous Robots}, vol.~38, no.~2, pp. 143--160, 2015.

\bibitem{hoff2000interactive}
K.~Hoff, T.~Culver, J.~Keyser, M.~C. Lin, and D.~Manocha, ``Interactive motion
  planning using hardware-accelerated computation of generalized voronoi
  diagrams,'' \emph{Proc. of the {IEEE} Intl. Conf. on Robot. and Autom.
  ({ICRA})}, vol.~3, pp. 2931--2937, 2000.

\bibitem{foskey2001voronoi}
M.~Foskey, M.~Garber, M.~C. Lin, and D.~Manocha, ``A voronoi-based hybrid
  motion planner,'' \emph{Proc. of the {IEEE/RSJ} Intl. Conf. on Intell. Robots
  and Syst.({IROS})}, vol.~1, pp. 55--60, 2001.

\bibitem{kavraki1996probabilistic}
L.~E. Kavraki, P.~Svestka, J.-C. Latombe, and M.~H. Overmars, ``Probabilistic
  roadmaps for path planning in high-dimensional configuration spaces,''
  \emph{IEEE transactions on Robotics and Automation}, vol.~12, no.~4, pp.
  566--580, 1996.

\bibitem{tagliasacchi2012mean}
A.~Tagliasacchi, I.~Alhashim, M.~Olson, and H.~Zhang, ``Mean curvature
  skeletons,'' in \emph{Computer Graphics Forum}, vol.~31, no.~5.\hskip 1em
  plus 0.5em minus 0.4em\relax Wiley Online Library, 2012, pp. 1735--1744.

\bibitem{tagliasacchi2009curve}
A.~Tagliasacchi, H.~Zhang, and D.~Cohen-Or, ``Curve skeleton extraction from
  incomplete point cloud,'' in \emph{ACM SIGGRAPH 2009 papers}, 2009, pp. 1--9.

\bibitem{tagliasacchi20163d}
A.~Tagliasacchi, T.~Delame, M.~Spagnuolo, N.~Amenta, and A.~Telea, ``3d
  skeletons: A state-of-the-art report,'' in \emph{Computer Graphics Forum},
  vol.~35, no.~2.\hskip 1em plus 0.5em minus 0.4em\relax Wiley Online Library,
  2016, pp. 573--597.

\bibitem{zivkovic2005hierarchical}
Z.~Zivkovic, B.~Bakker, and B.~Krose, ``Hierarchical map building using visual
  landmarks and geometric constraints,'' in \emph{2005 IEEE/RSJ International
  Conference on Intelligent Robots and Systems}.\hskip 1em plus 0.5em minus
  0.4em\relax IEEE, 2005, pp. 2480--2485.

\bibitem{blanco2006consistent}
J.-L. Blanco, J.~Gonzalez, and J.-A. Fernandez-Madrigal, ``Consistent
  observation grouping for generating metric-topological maps that improves
  robot localization,'' in \emph{Proceedings 2006 IEEE International Conference
  on Robotics and Automation, 2006. ICRA 2006.}\hskip 1em plus 0.5em minus
  0.4em\relax IEEE, 2006, pp. 818--823.

\bibitem{fraundorfer2007topological}
F.~Fraundorfer, C.~Engels, and D.~Nist{\'e}r, ``Topological mapping,
  localization and navigation using image collections,'' in \emph{2007 IEEE/RSJ
  International Conference on Intelligent Robots and Systems}.\hskip 1em plus
  0.5em minus 0.4em\relax Ieee, 2007, pp. 3872--3877.

\bibitem{vazquez2009spectral}
R.~Vazquez-Martin, P.~Nunez, A.~Bandera, and F.~Sandoval, ``Spectral clustering
  for feature-based metric maps partitioning in a hybrid mapping framework,''
  in \emph{2009 IEEE International Conference on Robotics and
  Automation}.\hskip 1em plus 0.5em minus 0.4em\relax IEEE, 2009, pp.
  4175--4181.

\bibitem{konolige2011navigation}
K.~Konolige, E.~Marder-Eppstein, and B.~Marthi, ``Navigation in hybrid
  metric-topological maps,'' in \emph{2011 IEEE International Conference on
  Robotics and Automation}.\hskip 1em plus 0.5em minus 0.4em\relax IEEE, 2011,
  pp. 3041--3047.

\bibitem{deits2015computing}
R.~Deits and R.~Tedrake, ``Computing large convex regions of obstacle-free
  space through semidefinite programming,'' in \emph{Algorithmic Foundations of
  Robotics XI}.\hskip 1em plus 0.5em minus 0.4em\relax Springer, 2015, vol.
  107, pp. 109--124.

\bibitem{liu2017planning}
S.~Liu, M.~Watterson, K.~Mohta, K.~Sun, S.~Bhattacharya, C.~J. Taylor, and
  V.~Kumar, ``Planning dynamically feasible trajectories for quadrotors using
  safe flight corridors in 3-d complex environments,'' \emph{IEEE Robotics and
  Automation Letters ({RA-L})}, pp. 1688--1695, 2017.

\bibitem{zhong2020generating}
X.~Zhong, Y.~Wu, D.~Wang, Q.~Wang, C.~Xu, and F.~Gao, ``Generating large convex
  polytopes directly on point clouds,'' \emph{arXiv preprint arXiv:2010.08744},
  2020.

\bibitem{chen2016online}
J.~Chen, T.~Liu, and S.~Shen, ``Online generation of collision-free
  trajectories for quadrotor flight in unknown cluttered environments,'' in
  \emph{2016 IEEE International Conference on Robotics and Automation
  (ICRA)}.\hskip 1em plus 0.5em minus 0.4em\relax IEEE, 2016, pp. 1476--1483.

\bibitem{gao2016online}
F.~Gao and S.~Shen, ``Online quadrotor trajectory generation and autonomous
  navigation on point clouds,'' in \emph{2016 IEEE International Symposium on
  Safety, Security, and Rescue Robotics (SSRR)}.\hskip 1em plus 0.5em minus
  0.4em\relax IEEE, 2016, pp. 139--146.

\bibitem{lin2021r3live}
J.~Lin and F.~Zhang, ``R3live: A robust, real-time, rgb-colored,
  lidar-inertial-visual tightly-coupled state estimation and mapping package,''
  \emph{arXiv preprint arXiv:2109.12400}, 2021.

\bibitem{zhou2021fuel}
B.~Zhou, Y.~Zhang, X.~Chen, and S.~Shen, ``Fuel: Fast uav exploration using
  incremental frontier structure and hierarchical planning,'' \emph{IEEE
  Robotics and Automation Letters}, vol.~6, no.~2, pp. 779--786, 2021.

\end{thebibliography}

\end{document}